\documentclass[11pt]{article}

\usepackage[preprint]{acl}

\usepackage{times}
\usepackage{latexsym}

\usepackage[T1]{fontenc}

\usepackage[utf8]{inputenc}

\usepackage{microtype}

\usepackage{inconsolata}
 
\usepackage{graphicx}
\graphicspath{{../}}
\usepackage{amsmath}
\usepackage{amssymb}
\usepackage{subcaption}
\usepackage{fancyhdr}

\usepackage{multirow}
\usepackage{booktabs}
\usepackage{makecell}
\usepackage{colortbl}
\usepackage{placeins}
\usepackage{subcaption} 
\usepackage{algorithm}      
\usepackage{algorithmic}    
\usepackage{xcolor}

\usepackage{tcolorbox}
\tcbuselibrary{skins, breakable}
\newtcolorbox{promptbox}[1]{
    enhanced,
    breakable,
    colback=gray!3,             
    colframe=black!70,          
    arc=2mm,                    
    boxrule=0.6pt,              
    top=4mm, bottom=4mm,
    left=4mm, right=4mm,
    attach boxed title to top left={yshift=-1.2mm},
    boxed title style={
        colback=black!80,
        colframe=black!80,
        arc=2mm,
        boxrule=0pt,
        top=2mm, bottom=2mm,
        left=4mm, right=4mm,
    },
    title={\textbf{#1}},
}

\newtcolorbox{contentbox}[1]{
    enhanced,
    breakable,
    colback=gray!3,             
    colframe=black!70,          
    arc=2mm,                    
    boxrule=0.6pt,              
    top=4mm, bottom=4mm,
    left=4mm, right=4mm,
}
\title{Mistake Notebook Learning: Batch-Clustered Failures for Training-Free Agent Adaptation}

\author{First Author \\
  Affiliation / Address line 1 \\
  Affiliation / Address line 2 \\
  Affiliation / Address line 3 \\
  \texttt{email@domain} \\\And
  Second Author \\
  Affiliation / Address line 1 \\
  Affiliation / Address line 2 \\
  Affiliation / Address line 3 \\
  \texttt{email@domain} \\}
  
\author{
Xuanbo Su\textsuperscript{1} \quad
Yingfang Zhang\textsuperscript{2} \quad
Hao Luo\textsuperscript{1} \quad
Xiaoteng Liu\textsuperscript{3} \quad
Leo Huang\textsuperscript{1}\\
\\
\textsuperscript{1}Bairong Inc., Beijing, China \\
\textsuperscript{2}School of Mathematics, Harbin Institute of Technology, Harbin, China \\
\textsuperscript{3}School of Software, Jilin University, Changchun, China \\
 \small{
   \textbf{Correspondence:} \href{huangling@brgroup.com}{huangling@brgroup.com}
 }
}

\begin{document}
\maketitle
\begin{abstract}
With the growing adoption of Large Language Model (LLM) agents in persistent, real-world roles, they naturally encounter continuous streams of tasks and inevitable failures. A key limitation, however, is their inability to systematically learn from these mistakes, forcing them to repeat identical errors in similar contexts. Unlike prior training-free methods that primarily store raw instance-level experience or focus on retrieving successful trajectories, we propose Mistake Notebook Learning (MNL), a novel memory framework that enables agents to self-curate generalizable guidance from batch-clustered failures. This mechanism allows agents to distill shared error patterns into structured ``mistake notes,'' updating an external memory only when batch performance improves to ensure stability. To further amplify adaptability, we integrate MNL with test-time scaling, leveraging aggregated failure patterns to actively steer the search process away from known pitfalls. Experiments on mathematical reasoning, Text-to-SQL, and interactive agent benchmarks show that MNL achieves competitive performance compared to existing memory mechanisms and in-context methods in both effectiveness and efficiency. These findings position structured mistake abstraction as a critical lever for robust agent evolution, enabling continuous improvement without the cost of parameter updates. The code is available at \url{https://github.com/Bairong-Xdynamics/MistakeNotebookLearning/tree/main}.
\end{abstract}
\section{Introduction}
Parameter-tuning is a standard approach for LLM adaptation but suffers from high computational costs, fragility to distribution shifts, and test-time rigidity in dynamic environments \citep{zeng2023agenttuning,chen2023fireact,zhai2025agentevolvertowardsefficient,Wang_2024}, hindering the rapid iteration essential for continual learning.

Training-free context methods offer an alternative, typically falling into two paradigms. Prompt-based optimization refines a single system prompt \citep{yang2024largelanguagemodelsoptimizers,zhou2023largelanguagemodelshumanlevel,pryzant2023automaticpromptoptimizationgradient} but often suffers from context length constraints and signal dilution. Memory-based approaches store instance-level experience \citep{shinn2023reflexionlanguageagentsverbal,zhao2024expelllmagentsexperiential,zhang2024incontextprinciplelearningmistakes} to correct errors locally. However, they frequently lack subject-level abstraction, resulting in brittle behavior with limited generalization.

We introduce Mistake Notebook Learning (MNL), a training-free memory framework where the Tuner Model clusters failures by subject within batches via prompted subject clustering, distills shared error patterns into structured guidance, and commits updates only when batch performance improves. MNL positions adaptation as memory construction and context curation rather than weight updates, integrating with test-time scaling (TTS) to steer search away from systematically erroneous paths.

Across diverse domains including mathematics, Text-to-SQL, and agentic tasks, MNL demonstrates significant improvements with concise prompts and compact memory structures. Our experiments indicate that converting mistakes into generalized guidance serves as an effective lever for robust, low-overhead adaptation, achieving competitive performance compared to parameter-tuning baselines while maintaining efficiency.

Our contributions are threefold: (1) A general framework that enables evolution via batch-clustered mistake abstraction and structured guidance memory. (2) A conservative accept-if-improves rule that stabilizes memory evolution and prevents regressions. (3) Comprehensive validation across diverse domains—including mathematical reasoning, Text-to-SQL, and agentic workflows—demonstrating MNL's effectiveness and its compatibility with test-time scaling strategies.
 
\section{Related Work}

\paragraph{Agent Evolution and Memory Systems}
Strategies for agent evolution are generally categorized into parameter-tuning and training-free paradigms. Training-based methods, such as \textit{AgentEvolver} \citep{zhai2025agentevolvertowardsefficient}, \textit{FireAct} \citep{chen2023fireact}, and \textit{AgentTuning} \citep{zeng2023agenttuning}, typically rely on computationally intensive pipelines involving Supervised Fine-Tuning (SFT), Reinforcement Learning (RL), or evolutionary optimization \citep{qiu2025evolution} to internalize capabilities into model weights. In contrast, Training-Free approaches leverage memory mechanisms to enable self-evolution without gradient updates. Memory modules have become a cornerstone in these systems, enabling agents to leverage historical context for enhanced decision-making \citep{Wang_2024}. Contemporary memory systems adopt diverse storage formats, ranging from unstructured textual logs \citep{park2023generativeagentsinteractivesimulacra} and latent vector embeddings to structured knowledge graphs. Recent advancements have further integrated Reinforcement Learning (RL) to optimize memory management policies \citep{yan2025memoryr1enhancinglargelanguage,wang2025memalpha}. For example, \textit{Agentic Context Engineering (ACE)} \citep{zhang2025agenticcontextengineeringevolving} treats context as an evolving ``playbook,'' employing modular generation and reflection. \textit{Memento} \citep{zhou2025mementofinetuningllmagents} reframes continual learning as memory-based online reinforcement learning, employing a Case-Based Reasoning (CBR) mechanism to update memory without altering model parameters. Similarly, \textit{Training-Free GRPO} \citep{cai2025trainingfreegrouprelativepolicy} leverages group-relative semantic advantages to distill experiential knowledge into prompt-based token priors.

\paragraph{Learning from Mistakes}
Learning from mistakes is a critical capability for intelligent systems. Early works like \textit{Reflexion} \citep{shinn2023reflexionlanguageagentsverbal} and \textit{Self-Refine} \citep{madaan2023selfrefineiterativerefinementselffeedback} utilize iterative verbal feedback to correct errors within a single session. However, these corrections are often transient and not retained for future tasks. To address this, recent research focuses on persistent learning. \textit{LEAP} \citep{an2024learningmistakesmakesllm} and \textit{CoTErrorSet} \citep{tong-etal-2024-llms} explicitly fine-tune models on error-correction pairs to internalize mistake-avoidance capabilities. In the context of in-context learning, \textit{ExpeL} \citep{zhao2024expelllmagentsexperiential} and \textit{In-Context Principle Learning} \citep{zhang2024incontextprinciplelearningmistakes} extract principles or rules from failures to guide future inference. While these methods demonstrate the value of negative feedback, they often treat mistakes as isolated instances or rely on static rule extraction.

\paragraph{Mistake Notebook Learning (MNL)}
Distinct from prior works that focus on retrieving successful trajectories or procedural workflows, MNL establishes a framework centered on \textit{systematic mistake analysis}. While methods like ACE and Memento often operate at the instance level, MNL introduces a \textit{batch-clustered} mechanism that aggregates errors to distill high-level, generalized insights, thereby reducing the variance associated with instance-specific corrections. Furthermore, we  explore the integration of memory with \textit{Test-Time Scaling (TTS)}. Unlike ReasoningBank \citep{ouyang2025reasoningbankscalingagent}, which enhances capabilities by retrieving successful reasoning traces, MNL synergizes its ``Mistake Notebook'' with TTS to actively mitigate potential errors. MNL demonstrates superior efficiency and adaptability in complex agentic workflows compared to vanilla scaling approaches.

\section{Methodology}
\label{sec:method}

\begin{figure*}[t]
\vskip 0.1in
\begin{center}
\centerline{\includegraphics[width=0.95\textwidth]{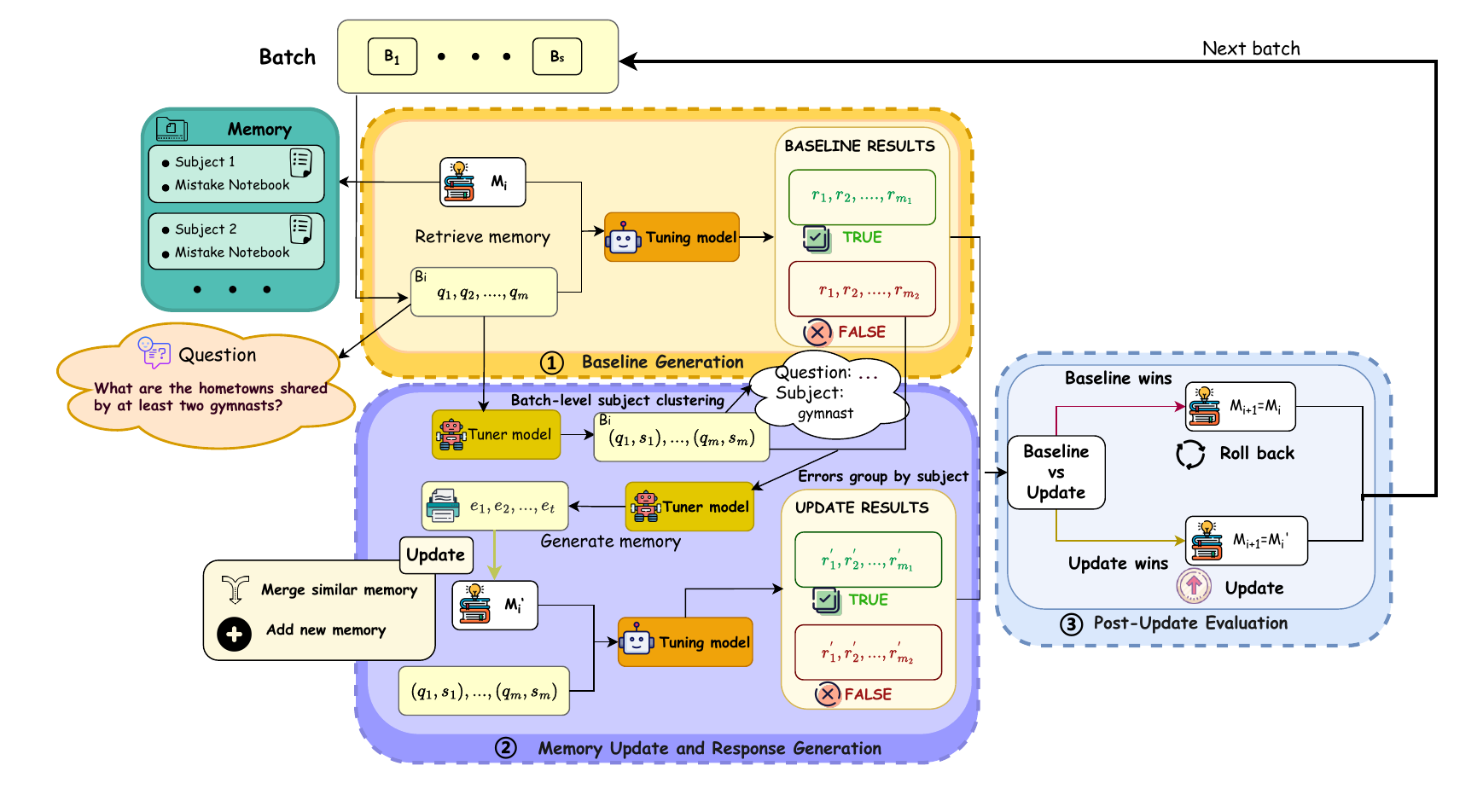}}
\caption{Overview of Mistake Notebook Learning (MNL). 
By utilizing a \textbf{Tuning Model} (the agent being improved) and a \textbf{Tuner Model} (the supervisor analyzing errors),
the whole process consists of three steps:
1) Baseline Generation — The Tuning Model produces initial responses with the current prompt and memory to establish a performance baseline.
2) Memory Update and Response Generation — The Tuner Model performs batch-level subject clustering via prompts, analyzes baseline errors, creates structured guidance items, and selectively updates the memory. The Tuning Model then generates updated responses.
3) Post-Update Evaluation — Compare performance before and after the update to assess the effectiveness of the revised memory and decide whether to accept the update.}
\label{fig:process}
\end{center}
\vskip -0.2in
\end{figure*}

\begin{figure}[t]
\centering

\begin{subfigure}{0.47\columnwidth}
    \centering
    \includegraphics[width=\textwidth]{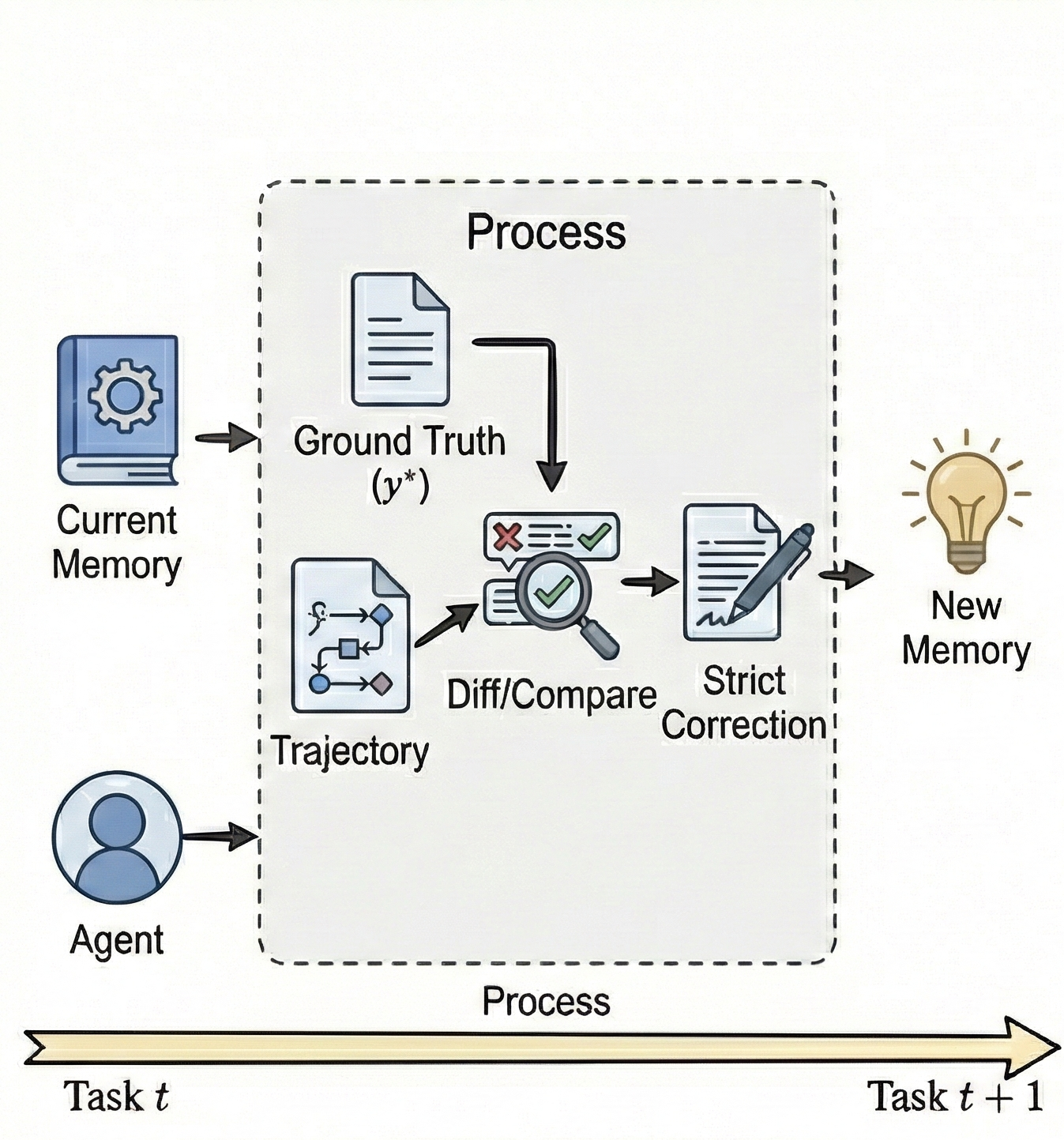}
    \caption{Supervised Evolution}
    \label{subfig:regimes_a}
\end{subfigure}
\hfill
\begin{subfigure}{0.5\columnwidth}
    \centering
    \includegraphics[width=\textwidth]{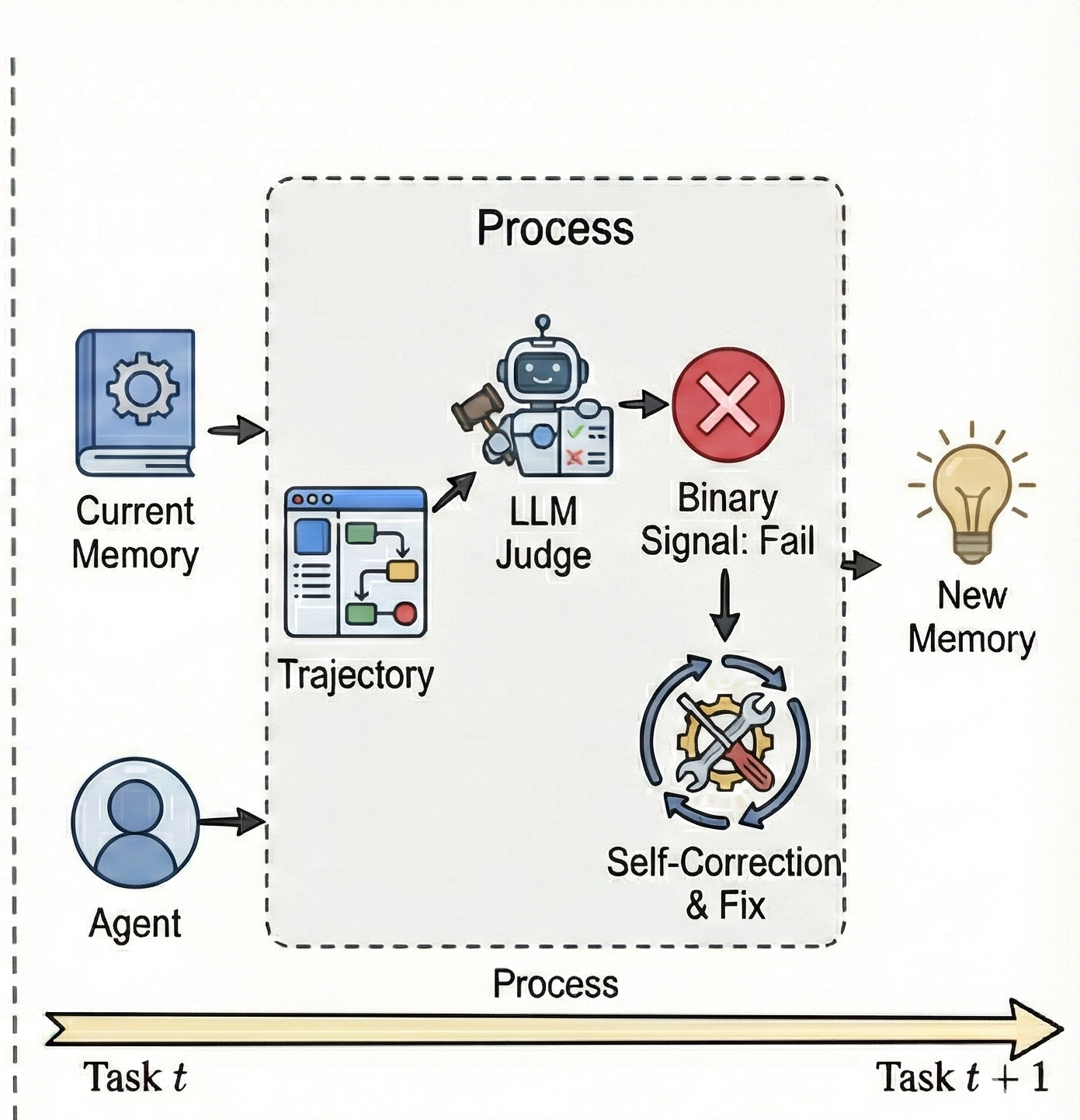}
    \caption{Self-Evolution}
    \label{subfig:regimes_b}
\end{subfigure}

\caption{Comparison of the two operating regimes in MNL. (a) \textbf{Supervised Evolution} relies on explicit ground truth ($y^*$) for direct trajectory correction. (b) \textbf{Self-Evolution} leverages a proxy verifier (LLM Judge) to generate binary utility signals, enabling the agent to evolve solely from interaction experience without accessing ground truth labels.}
\label{fig:regimes}
\end{figure}

\subsection{Method Overview}
We propose \textbf{Mistake Notebook Learning (MNL)}, a memory-based, training-free, self-evolving framework designed to enhance the problem-solving proficiency of LLM-based agents. As illustrated in Figure \ref{fig:process}, MNL operates with two distinct roles to enable evolution: the \textbf{Tuning Model} ($\pi_\theta$), which generates responses and whose performance we aim to improve; and the \textbf{Tuner Model} ($\pi_{\text{tuner}}$), which analyzes failures and updates the memory. At its core, MNL maintains and continuously refines an external dynamic memory $\mathcal{M}$. Unlike prior approaches that accumulate instance-level experiences \cite{zhou2025mementofinetuningllmagents, zheng2024synapse, wang2025agent}, MNL leverages a \textbf{batch-clustered} mechanism: failed trajectories are clustered under shared semantic subjects by the Tuner Model via prompts, and generalized error patterns and corrective strategies are distilled, forming stable and transferable memory \cite{zhang2024incontextprinciplelearningmistakes}. To ensure stability, updates are accepted only when they improve batch performance; otherwise, the previous memory state is retained. The framework follows a closed-loop process, iteratively performing baseline generation, memory update, and post-update evaluation to enable agents to self-evolve across different task domains and learning paradigms. The implementation details are presented in Appendix \ref{app:implementation_details}. The specific prompts utilized in this process are detailed in Appendix \ref{app:prompts}.

As illustrated in Figure~\ref{fig:regimes}, we distinguish two \emph{learning regimes}. MNL operates in a \emph{Supervised Evolution} regime (Figure~\ref{subfig:regimes_a}), where explicit ground-truth answers are used to determine output correctness and provide feedback for memory construction. For agent tasks, MNL operates in a \emph{Self-Evolution} regime (Figure~\ref{subfig:regimes_b}), in which a proxy verifier implemented as an LLM-based judge \cite{ouyang2025reasoningbankscalingagent,gu2025surveyllmasajudge, sun2025seagentselfevolvingcomputeruse} assesses trajectory outcomes and produces binary utility signals, which enables memory generation without access to ground-truth labels, with the specific LLM judge prompts provided in Appendix~\ref{app:llm_judge}. Furthermore, we combine MNL memory with Test-Time Scaling (TTS) in agent tasks, performing memory induction on the test set prior to the final evaluation.

\subsection{Problem Formulation}
\label{sec:problem_formulation}
We formulate MNL as a context optimization problem aiming at constructing a semantic memory \(\mathcal{M}\) that maximizes the expected reward of a frozen policy \(\pi_\theta\). Rather than updating model parameters, MNL improves performance by refining the memory \(\mathcal{M}\) to ensure that the retrieved context \(\text{Ret}(x, \mathcal{M})\) provides effective guidance for each input \(x\).

Formally, for a task distribution $\mathcal{D} = \{(x, y)\}$, we seek an optimal memory
\begin{align}
\mathcal{M}^* = \arg\max_{\mathcal{M}} \;&
\mathbb{E}_{(x, y) \sim \mathcal{D}}
\big[ R(\pi_\theta(z), y) \big], \label{eq:memory_objective} \\
\text{where } z \;=\;& x \oplus \text{Ret}(x, \mathcal{M}). \label{eq:context_definition}
\end{align}
where \(\oplus\) denotes prompt concatenation and \(R(\cdot)\) is a reward function. In supervised settings, \(R\) is derived from ground-truth labels; in self-correction settings, it is estimated by a proxy verifier.

The optimization of $\mathcal{M}$ is delegated to a dedicated \textbf{Tuner Model} ($\pi_{\text{tuner}}$). Distinct from the inference role of \textbf{Tuning Model} $\pi_\theta$, the Tuner Model acts as a reflective supervisor. It aggregates failed trajectories from $\pi_\theta$, performs prompted subject clustering to identify systematic error patterns across batches, and synthesizes structured corrections to update the memory. This decoupling of execution ($\pi_\theta$) and evolution ($\pi_{\text{tuner}}$) allows MNL to support various deployment configurations, such as self-correction \cite{madaan2023selfrefineiterativerefinementselffeedback} or expert-guided distillation \cite{kim2025guidingreasoningsmalllanguage}. Detailed prompts governing the Tuner Model's operations are provided in Appendix \ref{app:prompts}.

Depending on the available resources, these roles can be instantiated in two \emph{tuning configurations}: (1) \textbf{Self-Tuning}, where a single base model functions as both the Tuning Model and the Tuner Model to autonomously refine its own memory; and (2) \textbf{Cross-Model Tuning}, where a stronger model serves as the Tuner Model to distill high-quality guidance for a weaker Tuning Model. Table~\ref{tab:self_tuning} later compares these two configurations on Qwen3-8B.

\subsection{The MNL Evolution Protocol}
As illustrated in Figure \ref{fig:process}, the MNL framework operates through a closed-loop iterative process consisting of three sequential steps: Baseline Generation, Memory Update, and Post-Update Evaluation. This cycle ensures the continuous refinement of the memory $\mathcal{M}$ based on empirical performance feedback. 

\paragraph{Step 1: Baseline Generation} 
The process commences with the Tuning Model $\pi_\theta$ generating initial responses for a batch of queries. For each query $x$, the system retrieves relevant memory entries $\text{Ret}(x, \mathcal{M})$ to serve as advisory context. The Tuning Model is instructed to critically evaluate this context rather than blindly following it, thereby mitigating the risk of hallucination (see Appendix \ref{app:applicable}). These initial responses establish a performance baseline for the current iteration.

\paragraph{Step 2: Memory Update and Response Generation}
The Tuner Model $\pi_{\text{tuner}}$ analyzes the failed trajectories identified in the baseline generation. To solve the context optimization problem in Eq.~\eqref{eq:memory_objective}-\eqref{eq:context_definition}, we propose a batch-clustered approach that extracts generalized error patterns from failed trajectories. At iteration $t$, we sample a batch $\mathcal{B}=\{(x_i,y_i)\}_{i=1}^B\sim\mathcal{D}$ and generate baseline outputs $\hat y_i=\pi_\theta\big(x_i\oplus \text{Ret}(x_i,\mathcal{M}_t)\big)$. Define the failure index set $\mathcal{F}=\{i\mid R(\hat y_i,y_i)=0\}$ (or thresholded for real-valued rewards). A subject mapper $\sigma:\mathcal{X}\to\mathcal{S}$ is implemented by the Tuner Model, which performs semantic categorization of failed queries into precise subjects (e.g., combining domain, problem type, and solution method) via prompted clustering. This mapping, as detailed in Appendix~\ref{app:subject_clustering}, induces subject clusters $S_s=\{i\in\mathcal{F}\mid \sigma(x_i)=s\}$ over the failure set. The tuner then extracts cluster-level guidance
\begin{align}
g_s = \mathcal{E}\big(\{(x_i,y_i,\hat y_i)\}_{i\in S_s};\mathcal{M}_t\big), \quad s\in\mathcal{S}_\mathcal{F},
\end{align}
where $\mathcal{E}$ is the extraction operator that distills structured guidance from multiple failed trajectories within the same subject. 
and updates memory via
\begin{align}
\mathcal{M}_{t+1}=\text{Update}\big(\mathcal{M}_t,\{(s,g_s)\}_{s\in\mathcal{S}_\mathcal{F}}\big).
\end{align}
This batch-level abstraction is coupled with the accept-if-improves criterion in Eq.~\eqref{eq:net_improvement} to ensure stable memory evolution. New memory nodes are integrated either by merging with existing similar entries or by appending them as new nodes (see Appendix \ref{app:guidance_extraction} and \ref{app:guidance_merging}). Following this update, the Tuning Model generates refined responses conditioning on the updated memory.

\paragraph{Step 3: Post-Update Evaluation} 
To ensure the reliability of memory evolution, the system compares the performance of the updated responses against the baseline. Let $\Delta_{\mathcal{B}}$ denote the net improvement in batch accuracy:
\begin{equation}\label{eq:net_improvement}
\begin{split}
\Delta_{\mathcal{B}} =
\sum_{i=1}^{B}
\Big(
&\mathbb{I}[R(\hat{y}'_i) > R(\hat{y}_i)] \\
&- \mathbb{I}[R(\hat{y}'_i) < R(\hat{y}_i)]
\Big),
\end{split}
\end{equation}
where $\hat{y}_i$ and $\hat{y}'_i$ correspond to the outputs before and after the update, respectively. The memory update is accepted if and only if $\Delta_{\mathcal{B}} > 0$; otherwise, the previous memory state is retained. This ensures that only beneficial updates are kept, preserving the integrity of the ``Mistake Notebook''.

\section{Experiments}

In this section, we empirically validate the effectiveness of Mistake Notebook Learning (MNL) across three modalities: mathematical reasoning, Text-to-SQL, and interactive agents. We evaluate both task performance and efficiency, reporting memory size and inference-time guidance-token length alongside accuracy or success metrics. We further study sensitivity to key design choices (e.g., batch-level abstraction and training epochs) and compare MNL with supervised fine-tuning and cross-model tuning.

\begin{table*}[!t]
  \centering
  \caption{\label{tab:main_results}
    Main results on AIME 2024/2025 and KaggleDBQA. \textbf{Acc}: Pass@32 (AIME) / EA (KaggleDBQA). \textbf{Mem}: memory entries. \textbf{Len}: average guidance tokens (lower is better). Best in \textbf{bold}; ``-'' not applicable.
  }
  \begingroup
  \small
  \setlength{\tabcolsep}{3.5pt}
  \renewcommand{\arraystretch}{1.05}
  \begin{tabular}{@{}l|cccc|ccc@{}}
    \toprule
    \multirow{2}{*}{\textbf{Method}} &
    \multicolumn{4}{c|}{\textbf{AIME 2024 / 2025}} &
    \multicolumn{3}{c}{\textbf{KaggleDBQA}} \\
    \cmidrule(lr){2-5} \cmidrule(lr){6-8}
    & \textbf{Acc-24 (\%)} & \textbf{Acc-25 (\%)} & \textbf{Mem} & \textbf{Len} & \textbf{EA (\%)} & \textbf{Mem} & \textbf{Len} \\
    \hline
    \multicolumn{8}{c}{\textbf{Qwen3-8B}} \\
    Vanilla & 30\% & 23\% & - & - & 19\% & - & - \\
    TFGO & 23\% & 23\% & - & 703 & 22\% & - & \textbf{34} \\
    Memento & 20\% & 27\% & 100 & 3100 & 15\% & 87 & 530 \\
    ACE & 27\% & 10\% & 100 & 7355 & 22\% & 98 & 6289 \\
    MNL & \textbf{33\%} & \textbf{30\%} & 51 & \textbf{67} & \textbf{28\%} & 50 & 752 \\
    \hline
    \multicolumn{8}{c}{\textbf{DeepSeek-V3.2-Exp}} \\
    Vanilla & 87\% & 80\% & - & - & 24\% & - & - \\
    TFGO & \textbf{93\%} & \textbf{90\%} & - & 696 & 24\% & - & \textbf{100} \\
    ACE & 80\% & 67\% & 163 & 21318 & 54\% & 96 & 9406 \\
    Memento & - & - & - & - & 19\% & 87 & 1419 \\
    MNL & 90\% & 83\% & 9 & \textbf{60} & \textbf{64\%} & 54 & 514 \\
    \hline
    \multicolumn{8}{c}{\textbf{Qwen3-Max}} \\
    Vanilla & \textbf{93\%} & \textbf{96\%} & - & - & 40\% & - & - \\
    TFGO & 90\% & 90\% & - & 1452 & \textbf{47\%} & - & \textbf{125} \\
    Memento & - & - & - & - & \textbf{47\%} & 87 & 992 \\
    MNL & \textbf{93\%} & \textbf{96\%} & 10 & \textbf{0} & 46\% & 54 & 375 \\
    \bottomrule
  \end{tabular}
  \endgroup
\end{table*}

\subsection{Experimental Setup}
We evaluate MNL on three reasoning modalities: Mathematical (AIME 2024/2025 \citep{aime}, GSM8K \citep{cobbe2021gsm8k}), Text-to-SQL (KaggleDBQA \citep{lee-etal-2021-kaggledbqa}, Spider \citep{yu2019spiderlargescalehumanlabeleddataset}), and Interactive Agent (Mind2Web \citep{deng2023mind2webgeneralistagentweb}, AppWorld \citep{trivedi2024appworldcontrollableworldapps}). Specifically, AIME utilizes DAPO-100 as the training set, comprising 100 problems randomly sampled from the DAPO-Math-17K dataset \citep{yu2025dapoopensourcellmreinforcement}. Following \citet{cai2025trainingfreegrouprelativepolicy}, Spider and GSM8K adopt their respective standard training and test splits, consistent with their official configurations: Spider uses 7,000 training examples and 1,034 development samples for evaluation, while GSM8K uses 7,473 training and 1,319 test samples. On KaggleDBQA, we use the 87 provided examples for training and evaluate on the 185 test samples. We employ Qwen3-8B \citep{qwen3}, DeepSeek-V3.2-Exp \citep{deepseekai2024deepseekv32}, and Qwen3-Max \citep{qwen3max} as base models. Evaluation metrics include Pass@32 for AIME, execution accuracy (EA) for Text-to-SQL, as well as Task Success (TS) and Step Accuracy (SA) for agent benchmarks. Vanilla baselines follow standard prompting strategies per benchmark.\footnote{Math and Text-to-SQL use Chain-of-Thought prompting \citep{wei2023chainofthoughtpromptingelicitsreasoning}; Mind2Web uses few-shot prompting aligned with prior work; AppWorld uses ReAct-style prompting \citep{yao2023reactsynergizingreasoningacting}.} Unless otherwise specified, we adopt the \emph{Self-Tuning} configuration (the tuner shares the same base model as the tuning model); Table~\ref{tab:self_tuning} later compares this with \emph{Cross-Model Tuning}. Detailed settings, datasets, and baselines are provided in Appendix \ref{EXPERIMENT:dataset}.

\subsection{Main Results: Effectiveness and Efficiency}

We first present results on standard reasoning benchmarks (Math and Text-to-SQL) where MNL operates under \emph{Supervised Evolution}, followed by interactive agent tasks under \emph{Self-Evolution}.

\paragraph{Mathematical Reasoning Results} 
Table \ref{tab:main_results} reports AIME 2024/2025 results. MNL improves or preserves accuracy across vanilla models while keeping memory compact. On Qwen3-8B, MNL achieves 33.0\%/30.0\% using 51 memory entries and a guidance-token length of 66.8, outperforming retrieval-heavy baselines (e.g., Memento, ACE) that rely on much longer contexts (3k–7k tokens). On DeepSeek-V3.2-Exp, MNL attains 90.0\%/83.0\% with 9 memory entries and a guidance-token length of 60; TFGO achieves slightly higher AIME accuracy but requires longer traces with a length of 696 tokens. On Qwen3-Max, MNL matches the vanilla model on both years (93.0\%/96.0\%), indicating no degradation on a highly capable base model. On GSM8K (Table~\ref{tab:gsm8k_spider_sft}), MNL improves accuracy by +2.1 points and narrows the gap to SFT to 0.4 points.

\paragraph{Text-to-SQL Results} 
Table~\ref{tab:main_results} reports execution accuracy on KaggleDBQA. Across base models, MNL improves over vanilla while keeping memory compact and guidance-token length moderate. The gains are most pronounced on DeepSeek-V3.2-Exp: MNL boosts EA from 24.0\% to 64.0\% using 54 memory entries and 514 guidance tokens, whereas ACE \citep{zhang2025agenticcontextengineeringevolving} attains 54.0\% but requires a much longer context of 9406 tokens. On Qwen3-8B, MNL reaches 28.0\% with 752 guidance tokens, substantially shorter than ACE (6289 tokens) and far more accurate than vanilla (19.0\%). On Qwen3-Max, MNL improves over vanilla (46.0\% vs.\ 40.0\%) with a compact memory and a short prompt compared to retrieval-heavy alternatives like Memento \citep{zhou2025mementofinetuningllmagents}. Table~\ref{tab:gsm8k_spider_sft} further shows that MNL improves Spider accuracy over vanilla without parameter updates, narrowing the gap to SFT.

\paragraph{Interactive Agent Results} 
Tables \ref{tab:mind2web_results} and \ref{tab:appworld_results} show results on interactive agents under \emph{Self-Evolution}. 
The main accuracy metrics include Task Success (TS) and Step Accuracy (SA).
On Mind2Web, MNL improves Step Accuracy while reducing guidance-token length by orders of magnitude compared to retrieval/trajectory-heavy baselines (e.g., ACE \citep{zhang2025agenticcontextengineeringevolving} and Memento \citep{zhou2025mementofinetuningllmagents}). With DeepSeek-V3.2-Exp, MNL improves over vanilla on both Task Success and Step Accuracy (18.86/67.55 vs.\ 15.49/66.32) while using only 12 memory entries and 395 tokens; in contrast, ACE uses 58602 tokens. On AppWorld, MNL delivers low-overhead steering: on Qwen3-8B it improves Task Success (14.3 vs.\ 12.5) with 391 tokens, and on DeepSeek-V3.2-Exp it matches vanilla Task Success (73.2) while eliminating additional guidance tokens. Overall, MNL’s batch-level mistake abstraction yields compact guidance that remains effective for multi-step interactions without inflating the prompt.

\begin{table}[!t]
  \centering
  \small
  \setlength{\tabcolsep}{3pt}
  \caption{\label{tab:mind2web_results}
  Results on interactive agent tasks on Mind2Web (\%).}
  \begin{tabular}{@{}lcccc@{}}
    \toprule
    \textbf{Method} & \textbf{TS (\%)} & \textbf{SA (\%)} & \textbf{Mem} & \textbf{Len} \\
    \hline
    \rowcolor{gray!15}
    \multicolumn{5}{c}{\textbf{Qwen3-8B}} \\
    Vanilla model & 1.35\% & 11.54\% & - & - \\
    Memento & 0.00\% & 0.18\% & 1707 & 4749 \\
    ACE & 0.00\% & 0.00\% & 363 & 24284 \\
    MNL & \textbf{2.02\%} & \textbf{15.64\%} & 695 & \textbf{556} \\
    \hline
    \rowcolor{gray!15}
    \multicolumn{5}{c}{\textbf{DeepSeek-V3.2-Exp}} \\
    Vanilla model & 15.49\% & 66.32\% & - & - \\
    Memento & 0.34\% & 12.60\% & 1707 & 4822 \\
    ACE & 15.82\% & 57.80\% & 580 & 58602 \\
    MNL & \textbf{18.86\%} & \textbf{67.55\%} & 12 & \textbf{395} \\
    \hline
  \end{tabular}
\end{table}

\begin{table}[!t]
  \centering
  \small
  \setlength{\tabcolsep}{3pt}
  \caption{\label{tab:appworld_results}
    Results on interactive agent tasks on AppWorld (\%).}
  \begin{tabular}{@{}lccc@{}}
    \toprule
    \textbf{Method} & \textbf{TS (\%)} & \textbf{Mem} & \textbf{Len} \\
    \hline
    \rowcolor{gray!15}
    \multicolumn{4}{c}{\textbf{Qwen3-8B}} \\
    Vanilla model & 12.5\% & - & - \\
    Memento & 12.5\% & 50 & 707 \\
    ACE & 0.0\% & 61 & 3217 \\
    MNL & \textbf{14.3\%} & 12 & \textbf{391} \\
    \rowcolor{gray!15}
    \hline
    \multicolumn{4}{c}{\textbf{DeepSeek-V3.2-Exp}} \\
    Vanilla model & \textbf{73.2\%} & - & - \\
    Memento & 64.2\% & 56 & 602 \\
    ACE & 44.6\% & 8 & 6902 \\
    MNL & \textbf{73.2\%} & 0 & \textbf{0} \\
    \hline
  \end{tabular}
\end{table}

\paragraph{Integration with Test-Time Scaling (TTS)}
All main-table results use no-think base models. We additionally evaluate TTS-enabled variants (w/ think) in Tables \ref{tab:mind2web_tts} and \ref{tab:appworld_tts}. MNL remains compatible with TTS and provides consistent gains: As shown in Tables~\ref{tab:mind2web_tts}, on Mind2Web with a think-enabled Qwen3-8B model, Task Success improves from 1.01\% to 1.35\% and Step Accuracy from 11.13\% to 12.60\%; As shown in Tables~\ref{tab:appworld_tts}, on AppWorld with DeepSeek-Reasoner, Task Success improves from 75.0\% to 76.2\%.

\begin{table}[!t]
  \centering
  \small
  \setlength{\tabcolsep}{3pt}
  \caption{\label{tab:mind2web_tts}
  Results on Mind2Web with think-enabled (TTS) models and Mimo-v2 variants (\%).}
  \begin{tabular}{@{}lcccc@{}}
    \toprule
    \textbf{Method} & \textbf{TS (\%)} & \textbf{SA (\%)} & \textbf{Mem} & \textbf{Len} \\
    \hline
    \rowcolor{gray!15}
    \multicolumn{5}{c}{\textbf{Qwen3-8B-w/-think}} \\
    Vanilla model & 1.01\% & 11.13\% & - & - \\
    MNL & \textbf{1.35\%} & \textbf{12.60\%} & 695 & \textbf{505} \\
    \hline
    \rowcolor{gray!15}
    \multicolumn{5}{c}{\textbf{Mimo-v2-w/o-think}} \\
    Vanilla model & \textbf{8.75\%} & \textbf{41.12\%} & - & - \\
    MNL & 8.08\% & 40.71\% & 410 & \textbf{413} \\
    \hline
    \rowcolor{gray!15}
    \multicolumn{5}{c}{\textbf{Mimo-v2-w/-think}} \\
    Vanilla model & 10.77\% & 47.63\% & - & - \\
    MNL & \textbf{11.09\%} & \textbf{48.51\%} & 410 & \textbf{423} \\
    \hline
  \end{tabular}
\end{table}

\begin{table}[!t]
  \centering
  \small
  \setlength{\tabcolsep}{3pt}
  \caption{\label{tab:appworld_tts}
    Results on AppWorld with think-enabled (TTS) models and Mimo-v2 variants (\%).}
  \begin{tabular}{@{}lccc@{}}
    \toprule
    \textbf{Method} & \textbf{TS (\%)} & \textbf{Mem} & \textbf{Len} \\
    \hline
    \rowcolor{gray!15}
    \multicolumn{4}{c}{\textbf{Qwen3-8B-w/-think}} \\
    Vanilla model & 8.9\% & - & - \\
    MNL & \textbf{10.7\%} & 12 & \textbf{391} \\
    \rowcolor{gray!15}
    \hline
    \multicolumn{4}{c}{\textbf{DeepSeek-Reasoner}} \\
    Vanilla model & 75.0\% & - & - \\
    MNL & \textbf{76.2\%} & 4 & \textbf{341} \\
    \hline
    \multicolumn{4}{c}{\textbf{Mimo-v2-w/o-think}} \\
    Vanilla model & 69.6\% & - & - \\
    MNL & \textbf{71.4\%} & 1 & \textbf{206} \\
    \hline
  \end{tabular}
\end{table}

\subsection{Analysis}

\begin{figure}[t]
\centering
\includegraphics[width=\columnwidth]{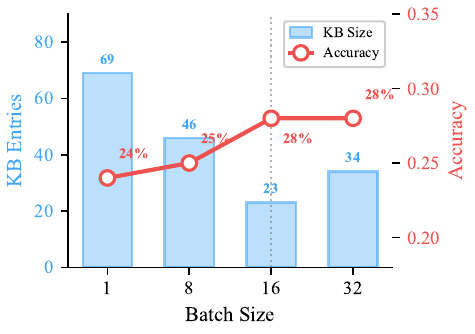}
\caption{Effect of batch size on KaggleDBQA. Batch size 16 achieves optimal balance: 28\% accuracy with only 23 KB entries vs.\ 24\% accuracy with 69 entries at batch size 1.}
\label{fig:batch_ablation}
\end{figure}

\paragraph{Ablation Study: Batch-Level Abstraction} 
Figure~\ref{fig:batch_ablation} confirms that batch-level abstraction reduces variance and improves generalization. Increasing batch size from 1 to 16 on KaggleDBQA improves accuracy by 17\% while reducing memory size by a factor of 3. This validates our hypothesis that aggregating errors allows the model to distill more general principles rather than overfitting to isolated instances. Intuitively, clustering semantically related failures and averaging their signals reduces estimation noise, leading to more reliable memory updates (see Appendix~\ref{app:batch_variance} for theoretical analysis).

\begin{figure}[t]
\centering
\includegraphics[width=\columnwidth]{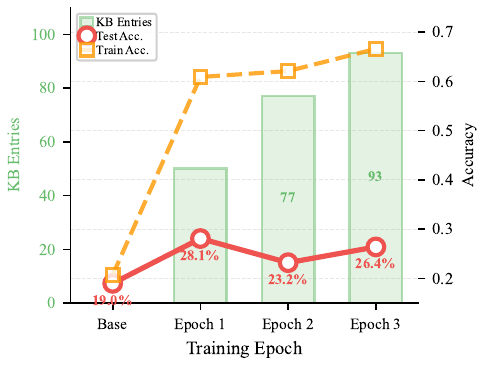}
\caption{Effect of training epochs on KaggleDBQA. Single-epoch achieves optimal test accuracy (28.1\%) with 50 KB entries. Multiple epochs cause cross-epoch overfitting: test accuracy drops to 23.2\% at epoch 2 while training accuracy rises to 62.1\%, demonstrating the memory overfits to training patterns.}
\label{fig:epoch_ablation}
\end{figure}

\paragraph{Ablation Study: Training Epochs} 
Figure~\ref{fig:epoch_ablation} shows that multi-epoch training leads to overfitting. Single-epoch training yields the highest test accuracy (28.1\%), while subsequent epochs increase training accuracy but degrade test performance. This suggests that the "Mistake Notebook" is best constructed by seeing each error type once and generalizing, rather than repeatedly fitting to the training set. We thus adopt single-epoch training as a standard practice.

\paragraph{Comparison with Supervised Fine-Tuning}
Figure~\ref{fig:sft_comparison} (in Appendix) compares MNL with SFT on Qwen3-8B. Table~\ref{tab:gsm8k_spider_sft} summarizes GSM8K and Spider results, showing that MNL narrows the gap to SFT on GSM8K and improves over the vanilla model on Spider without parameter updates.

\begin{table}[!t]
  \centering
  \normalsize
  \setlength{\tabcolsep}{6pt}
  \renewcommand{\arraystretch}{1.25}
  \caption{\label{tab:gsm8k_spider_sft}
  MNL vs.\ SFT on Qwen3-8B (Pass@1, \%). Best results in \textbf{bold}.}
  \begin{tabular}{@{}lccc@{}}
    \toprule
    \textbf{Dataset} & \textbf{Vanilla} & \textbf{MNL} & \textbf{SFT} \\
    \hline
    GSM8K & 91.8\% & 93.9\% & \textbf{94.3\%} \\
    Spider & 68.9\% & 71.7\% & \textbf{79.0\%} \\
    \hline
  \end{tabular}
\end{table}

\paragraph{Self-Tuning vs. Cross-Model Tuning}
We compare self-tuning (Qwen3-8B tuner) vs. cross-model tuning (DeepSeek-V3.2-Exp tuner) on Qwen3-8B. Table~\ref{tab:self_tuning} shows that while cross-model tuning yields slightly higher performance on KaggleDBQA (31.0\% vs. 28.0\%), self-tuning remains competitive. This confirms MNL's practical applicability even when a stronger supervisor model is not available.

\begin{table}[t]
  \centering
  \small
  \setlength{\tabcolsep}{4pt}
  \caption{\label{tab:self_tuning}Self-Tuning vs.\ Cross-Model Tuning on Qwen3-8B. Cross-Model Tuning (DeepSeek-V3.2-Exp as tuner) outperforms Self-Tuning, suggesting stronger tuner models can generate more effective guidance.}
  \begin{tabular}{@{}lcc@{}}
    \toprule
    \textbf{Dataset} &
    \textbf{Cross-Model} &
    \textbf{Self-Tuning} \\
    \hline
    AIME 2025 & 30.0\% & 30.0\% \\
    KaggleDBQA & \textbf{31.0\%} & 28.0\% \\
    \hline
  \end{tabular}
\end{table}

\paragraph{Training Cost Analysis}
Figure~\ref{fig:cost_abstract} highlights MNL's training-cost efficiency. On KaggleDBQA, MNL achieves 45.9\% accuracy at \$0.19 training cost, half the training cost of Memento. On GSM8K/Spider, MNL approaches SFT accuracy at $\sim$40\% lower training cost (see Appendix \ref{EXPERIMENT:COST}). This makes MNL particularly attractive for budget-constrained deployment.

\section{Conclusion}

In this paper, we introduced Mistake Notebook Learning (MNL), a training-free framework that shifts LLM adaptation from parameter updates to structured memory and context curation. By leveraging batch-wise error abstraction and a selective accept-if-improves rule, MNL evolves a compact memory that steers frozen LLM behavior without gradient computation.
We validated MNL under two regimes—Supervised Evolution with ground truth (Math and Text-to-SQL) and Self-Evolution with proxy judges (Mind2Web, AppWorld)—and observed consistent gains with short prompts and small memories. MNL is compatible with think-enabled models and enhances TTS performance, narrows the gap to SFT on GSM8K/Spider, and benefits from batch-level abstraction while avoiding multi-epoch overfitting. These results position memory- and context-centric adaptation as a practical, cost-efficient alternative to weight tuning for robust agent deployment.

\section*{Limitations}
\paragraph{Retrieval and Subject Granularity} MNL retrieves subject-level guidance via embedding similarity. Semantic asymmetry between concrete queries and abstract subjects can cause retrieval misses or mismatches, especially when subjects are overly broad or overly specific. Performance can therefore be sensitive to embedding quality, similarity thresholds, and the granularity of the subject taxonomy.

\paragraph{Feedback Quality and Verifier Reliability} In supervised settings, memory construction depends on the availability and correctness of ground-truth signals. In self-evolution settings, proxy verifiers such as LLM judges may introduce bias or inconsistency, which can propagate into the memory and lead to suboptimal or unstable updates. Although the accept-if-improves rule mitigates regressions at the batch level, it cannot fully eliminate systematic verifier errors.

\paragraph{Scalability, Maintenance, and Safety} As tasks and interactions grow, the memory can expand and increase retrieval and prompt-construction overhead. Additional mechanisms for memory consolidation and lifecycle management may be needed for long-running deployments. Finally, storing and reusing failure traces may raise privacy or safety concerns if trajectories contain sensitive information.

\bibliography{custom}

\appendix

\section{Appendix}

\label{app:batch_variance} 
\subsection{Why Batch-Level Abstraction Improves Decision Stability}

We provide a brief proof sketch explaining why batch-level (cluster-level) abstraction can reduce the probability of spurious updates under the \textit{accept-if-improves} criterion.

\paragraph{Setup.} Consider a fixed subject $s$ with cluster $S_s$. Let $\Delta_i$ denote the per-instance reward change induced by updating memory from $\mathcal{M}$ to $\mathcal{M}'$ with subject-level guidance:
\begin{equation}
\begin{split}
\Delta_i = &R\big(\pi_\theta(x_i\oplus\text{Ret}(x_i,\mathcal{M}')),y_i\big) \\
&-R\big(\pi_\theta(x_i\oplus\text{Ret}(x_i,\mathcal{M})),y_i\big).
\end{split}
\end{equation}

For theoretical intuition, we assume an additive model for $i\in S_s$:
\begin{equation}
\Delta_i = \mu_s + \varepsilon_i, \quad \mathbb{E}[\varepsilon_i]=0, \quad \mathrm{Var}(\varepsilon_i)=\sigma_s^2 < \infty.
\end{equation}
Here, $\mu_s$ captures the shared directional effect of the memory update on instances within the same semantic cluster, while $\varepsilon_i$ models instance-specific noise. We assume independent noise among cluster members, which is reasonable when instances are grouped by semantic similarity rather than arbitrarily.

\paragraph{Cluster-Average Estimator.} Define the cluster-average reward change:
\begin{equation}
\hat\mu_s = \frac{1}{|S_s|}\sum_{i\in S_s}\Delta_i.
\end{equation}
Standard results imply $\hat\mu_s$ is unbiased:
\[
\mathbb{E}[\hat\mu_s] = \mu_s,
\]
with variance
\[
\mathrm{Var}(\hat\mu_s) = \frac{\sigma_s^2}{|S_s|}.
\]

\paragraph{Implications for Accept-if-Improves.} The \textit{accept-if-improves} decision depends on the sign of the observed reward change. Let us consider the probability of an incorrect update decision given a true positive improvement $\mu_s > 0$:
\begin{align}
\text{One-by-one: } & \mathbb{P}(\Delta_i \le 0 \mid \mu_s > 0) = \mathbb{P}(\varepsilon_i \le -\mu_s), \\
\text{Cluster-avg: } & \mathbb{P}(\hat\mu_s \le 0 \mid \mu_s > 0) \nonumber \\
& \le 2 \exp\left(-c |S_s| \frac{\mu_s^2}{\sigma_s^2}\right),
\end{align}
where the second line follows from standard concentration inequalities for sub-Gaussian noise ($c>0$ is a constant). 

Thus, using the cluster-average $\hat\mu_s$ exponentially reduces the probability of spurious sign flips compared to one-by-one updates. In other words, batch-level abstraction directly improves the reliability of the \textit{accept-if-improves} decision rule.

\subsection{Implementation Details}
\label{app:implementation_details}
In this section, we provide a comprehensive overview of the MNL framework's implementation. We first define the structured Memory Schema ($\mathcal{M}$) in Appendix \ref{app:memory_schema}, designed to store actionable and generalizable insights efficiently. We then detail the technical execution of the MNL Evolution Protocol in Appendix \ref{app:implementation_details}, describing how the three-stage cycle of baseline generation, memory update, and post-update evaluation is realized in practice. Finally, we present the formal Algorithm that orchestrates this iterative MNL Evolution process in Appendix \ref{app:algorithm}. 

\subsubsection{Memory Schema and Storage}
\label{app:memory_schema}
To ensure scalability and efficient retrieval, we maintain the memory $\mathcal{M}$ in a JSONL format, where each entry is defined as a structured tuple $e = \langle s, g, \phi(s) \rangle$. The \textbf{Subject} ($s$) serves as a high-level semantic cluster identifier (e.g., ``SQL: Join conditions on null values'') to facilitate broad topic matching. The \textbf{Memory} ($g$) comprises five mandatory components to ensure actionability and safety: (1) \textit{Corrected Examples} that provide explicit mistake-answer pairs to ground the abstraction; (2) a \textit{Correct Approach} detailing the step-by-step reasoning methodology; (3) a \textit{Mistake Summary} identifying the root cause of the error; (4) a \textit{Generalizable Strategy} summarizing reusable problem-solving patterns; and (5) \textit{Anti-Patterns}, which are critical warnings specifying misapplication scenarios to prevent over-generalization. Finally, the \textbf{Embedding} $\phi(s)$ represents the semantic vector of the subject, pre-computed to enable efficient cross-modal retrieval against incoming query embeddings.

\subsubsection{MNL Evolution Protocol Implementation Details}
\label{app:mnl_protocol_impl}

\paragraph{Baseline Generation.}
The process commences with a retrieval-augmented generation step. For a batch of incoming queries, we compute query embeddings and perform a similarity search against the subject embeddings $\phi(s)$ in $\mathcal{M}$, retrieving the top-$k$ entries where the cosine similarity exceeds a specific threshold. These retrieved memory items are concatenated into the system context. To mitigate the risk of the Tuning Model $\pi_\theta$ blindly following potentially irrelevant historical advice, we append a specific applicability assessment instruction (see  Appendix ~\ref{app:applicable}). This compels the model to critically evaluate the relevance of the retrieved guidance before generating the initial baseline responses.

\paragraph{Memory Update and Response Generation.}
Following baseline generation, we employ a filtering mechanism to identify high-value learning opportunities. For domains with deterministic answers (e.g., Text-to-SQL, Math), correctness is determined by ground truth comparison; for open-ended agentic tasks, we utilize an LLM-as-a-judge to analyze the trajectory and produce binary success/failure signals (see Appendix \ref{app:llm_judge}). The update process operates at the subject level rather than the instance level. We first employ a \textit{Subject Clustering} step (prompt in Appendix~\ref{app:subject_clustering}) to group failed queries into semantic clusters. The Tuner Model $\pi_{\text{tuner}}$ then analyzes the collective failure trajectories within each cluster to distill the structured five-part memory described in Appendix~\ref{app:memory_schema}. To consolidate these insights into $\mathcal{M}$, we calculate the semantic similarity between the new subject and existing memory nodes. If the similarity exceeds a merge threshold, the new insights are fused into the existing node to refine the strategy; otherwise, a new node is appended.

\paragraph{Post-Update Evaluation.}
To guarantee the reliability of the evolving memory, we implement a closed-loop verification mechanism. The batch of queries is re-processed using the updated memory $\mathcal{M}'$, and we calculate the net performance improvement $\Delta_{\mathcal{B}}$ (see Eq.~\eqref{eq:net_improvement}). The memory update is committed only if $\Delta_{\mathcal{B}} > 0$; otherwise, the system rolls back to the previous state, ensuring that the memory $\mathcal{M}$ accumulates only beneficial and experimentally validated guidance.

\subsubsection{The MNL Evolution Algorithm}
\label{app:algorithm}
Algorithm~\ref{alg:mnl} presents the complete pseudocode for MNL, which consists of three main steps: (1) Baseline Generation, (2)  Memory Update and Response Generation, (3) Post-Update Evaluation.

\begin{algorithm}[!h]
\caption{Mistake Notebook Learning (MNL)}
\label{alg:mnl}
\begin{algorithmic}[1]
\small
\REQUIRE Task distribution $\mathcal{D}$, Tuning Model $\pi_\theta$, Tuner Model $\pi_{\text{tuner}}$, Reward function $R$, Batch size $B$
\STATE Initialize global memory $\mathcal{M} \leftarrow \emptyset$

\FOR{each batch $\mathcal{B} = \{(x_i, y_i)\}_{i=1}^B \sim \mathcal{D}$}
    
    \STATE \textbf{// Step 1: Baseline Generation}
    \STATE $\mathcal{Y}_{\text{base}} \leftarrow \emptyset$
    \FOR{$i = 1$ \TO $B$}
        \STATE $c_i \leftarrow \text{Ret}(x_i, \mathcal{M})$ 
        \STATE $z_i \leftarrow x_i \oplus c_i$
        \STATE $\hat{y}_i \leftarrow \pi_\theta(z_i)$ 
        \STATE $\mathcal{Y}_{\text{base}} \leftarrow \mathcal{Y}_{\text{base}} \cup \{\hat{y}_i\}$
    \ENDFOR

    \STATE \textbf{// Step 2: Memory Update and Response Generation}
    \STATE $\mathcal{F} \leftarrow \{ i \mid \hat{y}_i \text{ is identified as a failure} \}$ 
    
    \IF{$\mathcal{F} = \emptyset$}
        \STATE \textbf{continue} 
    \ENDIF

    \STATE $\mathcal{G}_{\text{fail}} \leftarrow \text{ClusterFailuresBySubject}(\{(x_i, \hat{y}_i)\}_{i \in \mathcal{F}}, \pi_{\text{tuner}})$
    \STATE $\mathcal{M}' \leftarrow \mathcal{M}$ \COMMENT{Initialize candidate memory}

    \FOR{each subject group $S \in \mathcal{G}_{\text{fail}}$}
        \STATE $\mathcal{P}_S \leftarrow \text{DistillPatternsAndStrategies}(S, \pi_{\text{tuner}})$ 

        \STATE $\mathcal{M}' \leftarrow \text{UpdateMemory}(\mathcal{M}', \mathcal{P}_S, \text{method}=\text{MergeOrAppend})$
    \ENDFOR

    \STATE $\mathcal{Y}_{\text{new}} \leftarrow \emptyset$
    \FOR{$i = 1$ \TO $B$}
        \STATE $\hat{y}'_i \leftarrow \pi_\theta(x_i \oplus \text{Ret}(x_i, \mathcal{M}'))$ 
        \STATE $\mathcal{Y}_{\text{new}} \leftarrow \mathcal{Y}_{\text{new}} \cup \{\hat{y}'_i\}$
    \ENDFOR

    \STATE \textbf{// Step 3: Post-Update Evaluation}
\STATE $\Delta_{\mathcal{B}} \leftarrow \sum_{i=1}^B \Big( \mathbb{I}[R(\hat{y}'_i) > R(\hat{y}_i)] - \mathbb{I}[R(\hat{y}'_i) < R(\hat{y}_i)] \Big)$
    \IF{$\Delta_{\mathcal{B}} > 0$}
        \STATE $\mathcal{M} \leftarrow \mathcal{M}'$ \COMMENT{Accept evolution}
    \ELSE
        \STATE Discard $\mathcal{M}'$ \COMMENT{Retain previous state}
    \ENDIF

\ENDFOR
\STATE \textbf{return} $\mathcal{M}$
\end{algorithmic}
\end{algorithm}

\subsection{Prompts Used in MNL Implementation}
\label{app:prompts}

\subsubsection{Applicability Assessment Prompt}
\label{app:applicable}
To prevent the model from blindly adopting retrieved memories that may be contextually mismatched, we prepend this instruction to the system prompt, enforcing a critical relevance check.
\begin{promptbox}{Applicability Assessment Prompt}

The following mistake notes are not necessarily tied to the current question, but you may use them to deepen your analytical approach.

\textbf{IMPORTANT}: Before applying any guidance below, carefully evaluate:
\begin{enumerate}
    \item Does the current problem match the applicability conditions stated in the guidance?
    \item Is the problem type and context similar to the examples in the guidance?
    \item If the problem is fundamentally different (e.g., combinatorics vs modulo arithmetic, complex numbers vs number theory), do NOT force-fit the guidance.
    \item Only use guidance that is clearly relevant to the current problem structure and requirements.
\end{enumerate}

Before solving, review the attached guidance. State whether it is: ``applicable'', ``partially applicable'', or ``irrelevant''. Use only applicable parts when answering.
    
\end{promptbox}

\subsubsection{Subject Clustering Prompt}
\label{app:subject_clustering}
We cluster each question into a high-specificity subject for RAG retrieval:

\begin{promptbox}{Subject Clustering Prompt}

You are an expert in categorizing questions into precise, high-relevance subjects for Retrieval-Augmented Generation (RAG).

Your goal is to assign each question a subject label that:
\begin{itemize}
    \item Maximizes retrieval relevance by precisely describing the problem type and solution method
    \item Groups only genuinely similar questions together (same domain AND same approach)
    \item Avoids over-broad categories that would match unrelated problems
\end{itemize}

Include: (1) Mathematical Domain, (2) Problem Type, (3) Solution Method.

Examples of GOOD subjects:
\begin{itemize}
    \item ``Combinatorics: Counting arrangements in grids with row and column sum constraints using stars and bars''
    \item ``Complex Analysis: Evaluating products over roots of unity using polynomial evaluation''
\end{itemize}

Examples of BAD subjects (too broad):
\begin{itemize}
    \item ``modulo arithmetic'' -- could match any modulo problem
    \item ``number theory'' -- could match any number theory problem
\end{itemize}
\end{promptbox}

\subsubsection{Structured Guidance Extraction Prompt}
\label{app:guidance_extraction}
We derive structured batch-level clustered memory with a prompt template designed to capture our five-component memory representation.
\begin{promptbox}{Structured Guidance Extraction Prompt}

You are an expert in analyzing model errors and maintaining a ``mistake notebook'' to improve future performance.

\textbf{Subject:} \{subject\}

\textbf{Error Examples:}

\{error\_context\}

\textbf{Task:} Extract insights from the mistakes and rewrite them as a structured mistake note.

Your response must include :

\textbf{1. Corrected Examples with mistake answers }
\begin{itemize}
    \item For each, include:
    \begin{itemize}
        \item The original question and mistake answer
        \item Correct answer and correct reasoning process
    \end{itemize}
\end{itemize}

\textbf{2. Correct Approach}
\begin{itemize}
    \item Provide the correct reasoning method or step-by-step approach that should be applied.
\end{itemize}

\textbf{3. Mistake Summary}
\begin{itemize}
    \item Identify the root cause behind the errors (reasoning flaw, misunderstanding of concept, missing steps, incorrect logic, etc.).
\end{itemize}

\textbf{4. Generalizable Strategy}
\begin{itemize}
    \item Summarize reusable problem-solving patterns and how to avoid future mistakes.
\end{itemize}

\textbf{5. ANTI-PATTERNS}

List specific things to AVOID:
\begin{itemize}
    \item Common ways this guidance gets misapplied
    \item Situations where following this guidance would be WRONG
    \item Red flags that indicate the guidance doesn't fit
\end{itemize}

Output format should resemble a mistake notebook entry: concise, structured, knowledge-focused, and reusable for similar future questions.
\end{promptbox}

\subsubsection{RAG-Based Guidance Merging Prompt}
\label{app:guidance_merging}
We merge new memory with related existing entries to enable memory updating:
\begin{promptbox}{RAG-Based Guidance Merging Prompt}

You are synthesizing guidance for subject: \{subject\}

\textbf{Existing guidance from related subjects in the memory:}

\{existing\_guidance\}

\textbf{New guidance to incorporate:}

\{new\_guidance\}

\textbf{Task:} Merge these into a single coherent guidance that:
\begin{itemize}
    \item Combines insights from related subjects with new guidance
    \item Eliminates redundancy while preserving key information and examples of the mistakes
    \item \textbf{Preserves and emphasizes applicability conditions}---clearly state when each method applies
    \item Focuses on actionable advice
    \item Maintains consistent style
    \item \textbf{Includes warnings about when NOT to apply the guidance} to avoid misapplication
\end{itemize}

\textbf{Merged guidance:}

\end{promptbox}

\subsection{LLM Judge Prompts}
\label{app:llm_judge}

To evaluate agent performance across different environments without relying solely on ground truth, we design specialized LLM-as-a-judge prompts. We tailor these prompts to the specific granularities of the Mind2Web and AppWorld benchmarks.

\subsubsection{Mind2Web Evaluation Prompts}
For the Mind2Web benchmark, we employ two distinct judging mechanisms. The Pairwise Comparison Judge (Appendix \ref{app:comparison_judge}) is utilized when the agent generates multiple candidate actions; it analyzes two options simultaneously to identify the optimal next step based on UI logic. Conversely, the Single Trajectory Judge (Appendix \ref{app:single_judge}) acts as a binary verifier, analyzing a specific action in isolation to determine its validity within the interaction flow.

\begin{promptbox}{Pairwise Comparison Judge Prompt}
\label{app:comparison_judge}
You are an expert in web navigation and user interface interaction.

Given this web navigation task: \{question\}

\textbf{Compare these two proposed actions and determine which one is MORE CORRECT:}

Action A: \{answer1\} 
Action B: \{answer2\}

\textbf{Evaluation criteria (in order of importance):}

Task relevance - Does this action directly help achieve the stated goal?

UI logic - Is this a logical next step given the current page state?

Element availability - Does the target element actually exist on the page?

Efficiency - Is this the most direct path to accomplish the task?

Think step by step, then respond with exactly ONE of these options:

"Action A is more correct"

"Action B is more correct"

"Both are equally correct or equally wrong"

Your response must start with one of these exact phrases. 
\end{promptbox}

\begin{promptbox}{Single Trajectory Judge Prompt }
\label{app:single_judge}
You are an expert in web navigation and user interface interaction evaluation. Your task is to determine if a candidate answer is correct for a given web navigation task.

You DO NOT have access to a ground truth answer, so you must judge strictly based on the provided web context (HTML), the user's task goal, and the interaction history.

\textbf{Context and Task:} \{question\}

\textbf{Proposed Action to Evaluate:} \{\text{candidate\_answer}\}

\textbf{Evaluation Steps:}

Goal Analysis: What is the user trying to achieve?

State Analysis: Based on previous actions, where are we in the flow?

Element Verification: Does the element selected in the proposed action exist in the HTML? Is it the correct element to interact with?

Action Validity: Is the action (CLICK, TYPE, SELECT) appropriate for this element and goal?

\textbf{Judgment Criteria:}

CORRECT: The action is the logical, necessary, and correct next step to advance the task.

INCORRECT: The action is irrelevant, interacts with the wrong element, uses the wrong action type, or hinders the task.

Respond with exactly ONE of the following lines, followed by your reasoning:

"Judgment: CORRECT"

"Judgment: INCORRECT" 
\end{promptbox}

\subsubsection{AppWorld Evaluation Prompt}
Unlike the step-by-step UI interactions in Mind2Web, the AppWorld benchmark requires evaluating complete API interaction chains and Python code execution. Therefore, the AppWorld Judge (Appendix \ref{app:appworld_prompt}) is designed to assess the full execution log, determining success based on the final output validity and the absence of fatal runtime errors.

\begin{promptbox}{AppWorld Execution Judge Prompt}
\label{app:appworld_prompt}
You are an expert Judge for an AI Agent execution log.
Your goal is to determine if the Agent successfully completed the user's task based on the provided execution log.

**Input Data:**

1. Task Question: "\{question\}"

2. Execution Log (JSON format):

<<<EXECUTION\_LOG\_START>>>

\{trajectory\}

<<<EXECUTION\_LOG\_END>>>

**Judgment Criteria:**

- **SUCCESS**: The agent found the requested information, performed the requested action, or provided the correct answer in the final turn. The code executed without fatal errors.

- **FAILURE**: The agent encountered a Python exception that stopped progress, got stuck in a loop, failed to call the correct API, authorized incorrectly, or the log ends abruptly without an answer.

**Output Format:**

You must output a single JSON object with the following structure:

{{
  "reasoning": "Brief analysis of why it succeeded or failed (max 50 words).",
  "is\_success": true or false
}}
\end{promptbox}

\subsection{EXPERIMENT DETAILS}
\subsubsection{Evaluation Protocol and Experimental Settings}\label{EXPERIMENT:dataset}

\paragraph{Evaluation Protocol.} We evaluate all settings under greedy decoding with \textbf{temperature} set to 0.0. We adopt task-dependent Pass@k: for \textbf{mathematical reasoning}, we report \textbf{Pass@32} on \textbf{AIME 2024/2025} to mitigate sampling variance on challenging problems, while all other tasks (including GSM8K, Text-to-SQL, and agent benchmarks) are evaluated with \textbf{Pass@1}. 

Evaluation metrics are task-specific: for \textbf{Text-to-SQL}, we report \emph{execution accuracy}, where a predicted SQL query is executed on the target database and matched against the gold execution result; for \textbf{mathematical reasoning}, we use \emph{normalized exact match} after symbolic simplification; for \textbf{agent tasks}, we evaluate \textbf{Mind2Web} using \emph{Task Success} and \emph{Step Accuracy}, and \textbf{AppWorld} by whether the agent successfully completes the user-defined \emph{task goal} in the real application environment. This “task-specific metrics” framing follows common practice in agent evaluation setups. 

\paragraph{Detailed Dataset Statistics and Splits.}
\textbf{AIME 2024/2025} training uses 100 examples randomly sampled from DAPO-Math-17K, and the test sets contain 30 problems each for AIME 2024 and AIME 2025. \textbf{GSM8K} uses 7,473 training examples and 1,319 test examples. \textbf{KaggleDBQA} uses 87 training examples and 185 test examples. Spider uses 7,000 training examples and 1,034 test examples. For \textbf{Mind2Web}, we select 3 subjects; for each subject we randomly sample 100 tasks. Each task contains average 6 steps, resulting in 1,707 training instances and 1,707 test instances. For \textbf{AppWorld}, we randomly sample one instance from each of the 56 scenarios for both training and testing, yielding 56 training instances and 56 test instances.

\paragraph{Table Conventions.} In all result tables, higher accuracy/success is better; best performance values are marked in \textbf{bold}. \textbf{Mem Cnt} denotes the number of memory entries constructed during training, and \textbf{Avg. Len (tok)} denotes the average number of guidance tokens used during inference. \textbf{Mem Cnt} is not applicable to TFGO due to prompt-based optimization. For cost reasons, we omit ACE results with Qwen3-Max where applicable; similarly, Memento and TFGO are omitted in some large-scale experiments due to their high token consumption and associated budgetary constraints.

\paragraph{Main Result Notes.} On AIME 2024/2025, MNL improves or preserves accuracy across base models with small memory. On KaggleDBQA, MNL yields consistent accuracy gains over the vanilla model and avoids the large context overhead of retrieval-heavy baselines. On Mind2Web and AppWorld, MNL improves agent success under self-correction while keeping inference context short.

Following our ablation results, all experiments use \textbf{single-epoch} training to avoid cross-epoch overfitting. Unless otherwise specified, we adopt the \textbf{Self-Tuning} setting, where the tuner shares the same architecture as the model being tuned. We use model-specific maximum generation lengths: \textbf{Qwen3-8B} and \textbf{Qwen3-Max} use a 32K-token limit, while \textbf{DeepSeek-V3.2-Exp} supports up to 8K tokens and is therefore evaluated with an 8K limit. All vanilla models are evaluated under the \textbf{no-think} setting.

\paragraph{Implementation Details.} For \textbf{Text-to-SQL}, \textbf{mathematical reasoning}, we use an \textbf{supervised evolution} setting, i.e., ground-truth answers are available during training/tuning to support explicit error attribution and feedback construction. In contrast, for \textbf{Mind2Web} and \textbf{AppWorld} we use an \textbf{self-evolution} setting: no ground-truth answers are provided during training, and an \textbf{LLM Judge} determines the agent’s \emph{task success} based on the final interaction outcome, which is then used to generate feedback signals for self-tuning. Finally, all methods (including TFGO, ACE, Memento, and MNL) are evaluated under the same protocol with ground true to ensure fair and reproducible comparison. 

\paragraph{Reproducibility Settings.}To ensure reproducibility, we fix the hyperparameters to Temperature=0, Presence Penalty=1.5, and Random Seed=42. For Qwen3-8B, we use max-tokens=32K; for DeepSeek and Qwen3-Max, we set max-tokens=8K. To better match different evaluation settings, we further set max-tokens=8K for Text-to-SQL and all ablation studies. For method-specific configurations, \textbf{MNL} uses epoch=1 and batch-size=16, with bge-m3 as the embedding model; during retrieval we set topk=1 and retrieval-threshold=0.6. \textbf{Memento} sets memory-max-pos-examples=4, memory-max-neg-examples=4, and memory-max-length=256, and also uses bge-m3 as the embedding model. For Memento, META-MODEL, EXEC-MODEL, and JUDGE-MODEL share the same backbone model. \textbf{TFGO} uses batchsize=64, rollout-concurrency=5, and rollout-max-tokens=4096. ACE uses epoch=1, max-num-rounds=3, and playbook-token-budget=80K; generator-model, reflector-model, and curator-model are instantiated with the same backbone model.The experimental settings for the other methods largely follow the default configurations provided in their open-source implementations.

\subsubsection{Cost Calculation Details}\label{EXPERIMENT:COST}
For KaggleDBQA, we use Qwen3-Max as the self-tuning mode and compute learning cost based on its official API pricing. Specifically, the pricing for Qwen3-Max-3 is 0.0032 RMB per 1k input tokens and 0.0128 RMB per 1k output tokens. The total learning cost is obtained by aggregating the number of input and output tokens generated during training according to these rates. For GSM8K and Spider, we use Qwen3-8B as the base model and conduct training on a single H20 GPU (141GB). The GPU usage cost is computed at a rate of \$3.99 per hour. GPU prices follow the publicly listed on-demand pricing at the time of experiments. MNL+Qwen3-8B completes training in 15 minutes on GSM8K, resulting in a total learning cost of \$0.99, while SFT+Qwen3-8B requires 30 minutes of training under the same hardware configuration, incurring a cost of \$1.98. On Spider, MNL+Qwen3-8B completes training in 30 minutes with a cost of \$1.98, whereas SFT+Qwen3-8B requires 50 minutes of training time and incurs a cost of \$3.32 under identical computational resources. All reported costs account for training only and exclude inference or evaluation overhead.

\begin{figure*}[t]
\centering
\includegraphics[width=0.9\textwidth,height=0.28\textheight,keepaspectratio]{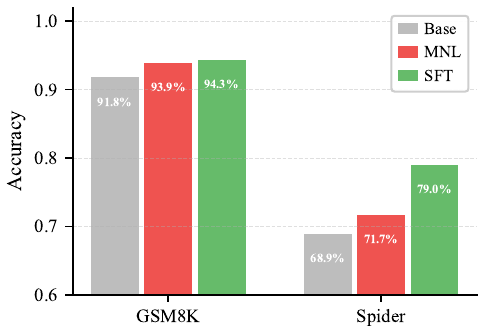}
\caption{MNL vs.\ SFT on Qwen3-8B. On GSM8K, MNL (93.9\%) nearly matches SFT (94.3\%). On Spider, SFT (79.0\%) leads, but MNL (71.7\%) improves over Vanilla model (68.9\%) without parameter updates.}
\label{fig:sft_comparison}
\end{figure*}

\begin{figure*}[t]
\centering
\includegraphics[width=0.85\textwidth,height=0.45\textheight,keepaspectratio]{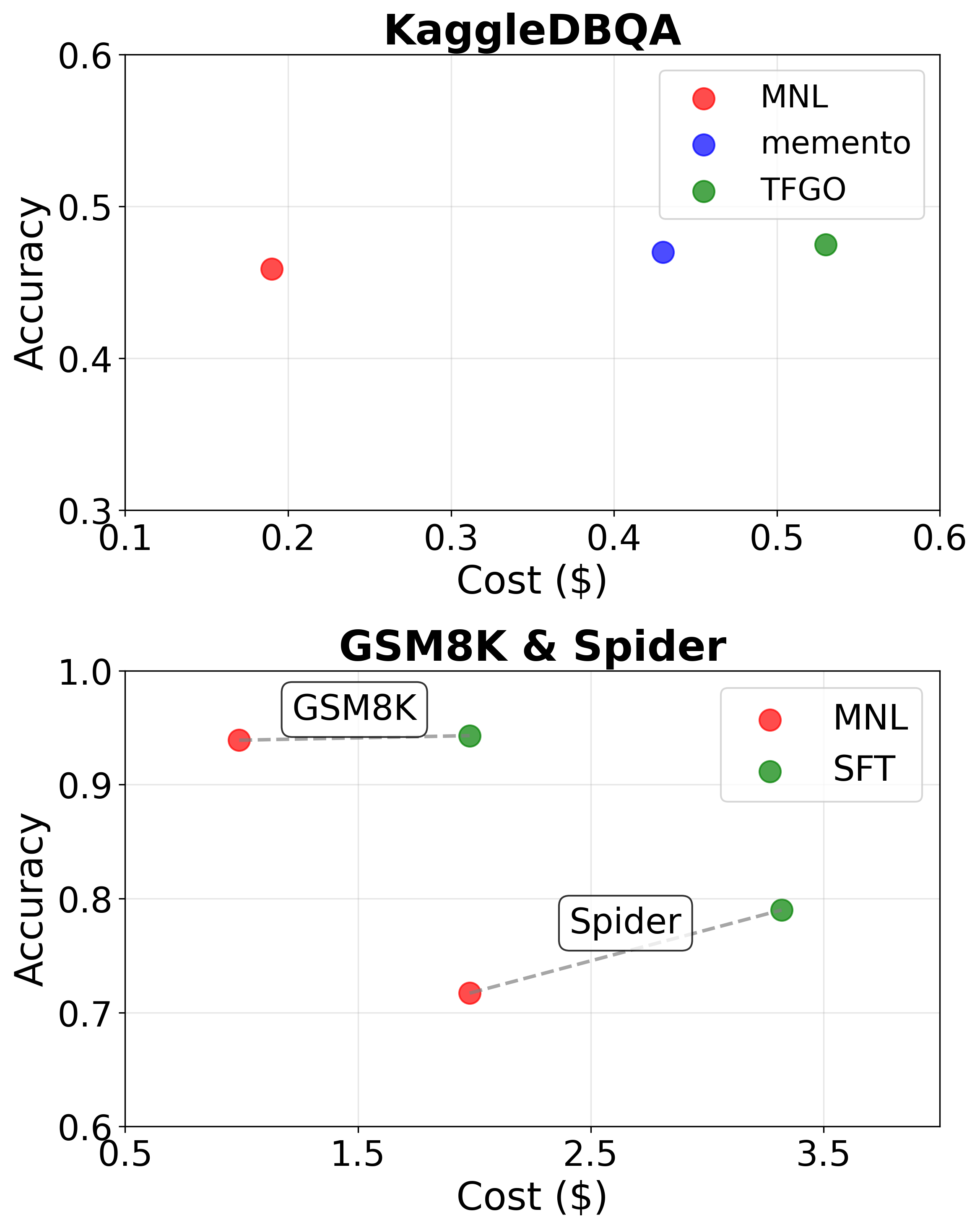}
\caption{Cost-accuracy trade-off. Top: On KaggleDBQA, MNL achieves 45.9\% accuracy at \$0.19, while Memento reaches 47.0\% accuracy at \$0.43 (2.3$\times$ cost). Bottom: On GSM8K/Spider, MNL approaches SFT accuracy at 40\% lower cost.}
\label{fig:cost_abstract}
\end{figure*}

\end{document}